\documentclass[10pt, conference, compsocconf]{IEEEtran}
\IEEEoverridecommandlockouts

\ifCLASSINFOpdf
\else
\fi
\usepackage{algorithmic,algorithm}

\usepackage{psfrag}
\usepackage{cite}
\usepackage[utf8]{inputenc}
\usepackage{multirow}
\usepackage{caption}
\usepackage{amssymb}
\usepackage{bm}
\usepackage{psfrag}
\usepackage{cite}
\usepackage{graphics}
\usepackage{fancyhdr}
\usepackage{eucal}
\usepackage[english]{babel}
\usepackage{ifpdf}
\usepackage{setspace}
\usepackage{float}
\usepackage[T1]{fontenc}
\usepackage{subfigure}
\usepackage{dsfont}
\usepackage{graphicx, framed}
\usepackage{textcomp}
\usepackage{tikz}
\usepackage{tikz-inet}
\usetikzlibrary{shapes,snakes,arrows,automata}

\newcommand{\ben}{\begin{enumerate}}
\newcommand{\een}{\end{enumerate}}

\newcommand{\bc}{\begin{center}}
\newcommand{\ec}{\end{center}}

\newcommand{\bit}{\begin{itemize}}
\newcommand{\eit}{\end{itemize}}

\newcommand{\ds}{\displaystyle}
\newcommand{\beq}{\begin{equation}}
\newcommand{\eeq}{\end{equation}}

\newcommand{\wre}{\mathbf{w}^{\rm{r}}}
\newcommand{\wi}{\mathbf{w}^{\rm{in}}}
\newcommand{\wo}{\mathbf{w}^{\rm{out}}}

\newcommand{\va}{\mathbf{a}}
\newcommand{\vb}{\mathbf{b}}

\newcommand{\x}{\mathbf{x}}
\newcommand{\y}{\mathbf{y}}
\newcommand{\ve}{\mathbf{v}}
\newcommand{\Na}{N_{\rm{a}}}
\newcommand{\Nb}{N_{\rm{b}}}
\newcommand{\Nx}{N_{\rm{x}}}

\newcommand{\p}{\mathbf{\mathbf{p}}}

\newcommand{\R}{\mathds{R}}

\newif\ifnotes\notestrue

\def\hgr#1{}

\def\hgr#1{}

\newcommand{\wmi}[1]{w^{\rm{#1}}_{mi}}
\begin{document}
%
\title{An Experimental Analysis of the Echo State Network Initialization Using the Particle Swarm Optimization
\thanks{This work was supported within the framework of the IT4Innovations Centre of Excellence project, reg. no. CZ.1.05/1.1.00/02.0070 supported by Operational Programme 'Research and Development for Innovations' funded by Structural Funds of the European Union and state budget of the Czech Republic and this article has been elaborated in the framework of the project New creative teams in priorities of scientific research, reg. no. CZ.1.07/2.3.00/30.0055. Additionally, this research is partially funded by the Spanish Ministry of Economy and
Competitiveness and FEDER within the roadME project TIN2011-28194 (http://roadme.lcc.uma.es) and the UMA-OTRI contract 8.06/5.47.4142 in
collaboration with the VSB-Technical University of Ostrava.}
}




%
\author{\IEEEauthorblockN{Sebasti\'an Basterrech\IEEEauthorrefmark{1},
Enrique Alba\IEEEauthorrefmark{1}\IEEEauthorrefmark{2},
V\'aclav Sn\'a\v{s}el\IEEEauthorrefmark{1}, 
}
\IEEEauthorblockA{\IEEEauthorrefmark{1}IT4Innovations\\
V\v{S}B-Technical University of Ostrava\\
Ostrava-Poruba, Czech Republic\\
Email: \{Sebastian.Basterrech.Tiscordio,Vaclav.Snasel\}@vsb.cz}
\IEEEauthorblockA{\IEEEauthorrefmark{2}\textit{Dept. Lenguajes y Ciencias de la Computaci\'on}\\
\textit{Universidad de M\'alaga}\\
M\'alaga, Spain\\
Email: eat@lcc.uma.es}
}


\maketitle

\begin{abstract}
This article introduces a robust hybrid method for solving supervised learning tasks, which uses the \textit{Echo State Network (ESN)} model and the \textit{Particle Swarm Optimization (PSO)} algorithm.
An ESN is a Recurrent Neural Network with the hidden-hidden weights fixed in the learning process.
The recurrent part of the network stores the input information in internal states of the network.
Another structure forms a free-memory method used as supervised learning tool.
The setting procedure for initializing the recurrent structure of the ESN model can impact on the model performance.
On the other hand, the PSO has been shown to be a successful technique for finding optimal points in complex spaces.
Here, we present an approach to use the PSO for finding some initial hidden-hidden weights of the ESN model.
We present empirical results that compare the canonical ESN model with this hybrid method on a wide range of benchmark problems.
\end{abstract}


%
\IEEEpeerreviewmaketitle

\section{Introduction}
A \textit{Recurrent Neural Network (RNN)} is a powerful tool for time-series modeling~\cite{Jaeger09}.
It has been used for solving supervised temporal learning tasks as well as for information processing in biological neural systems~\cite{Jaeger09,Maass02}. 
The recurrent topology of the network ensures that a non-linear transformation of the input information can be stored in internal states~\cite{Jaeger09}. 
In spite of that, recurrent networks present some limitations for solving real-world applications~\cite{Jaeger09}.
They can present high computational costs during the training process when a $1$st-order learning algorithm is used (for instance: gradient descent algorithm type)~\cite{Bengio94}.
During the~$90$s much effort was devoted to identify the learning problems of the RNNs.

At the beginning of the~$2000$s two models were introduced for designing and training RNNs.
They were independently developed and named \textit{Echo State Network (ESN)}~\cite{Jaeger01} and \textit{Liquid State Machines}~\cite{Maass02}. 
Since 2007 this trend has started to be popularly known under the name of~\textit{Reservoir Computing (RC)}~\cite{Schrauwen07}.
The RC approach is an attempt to resolve the limitations in the training, which overcome the limitations of convergence time.
A RC model is a RNN with the particularity that the weights involved in cyclic connections are deemed fixed during the training process.
%
The recurrent structure of the network is called~\textit{reservoir} and it is composed by the hidden-hidden weights. 
Another structure of the model called~\textit{readout} refers to the weight connections free of recurrences in the network, in graph terms the readout is composed by the free-circuit weights. 
Only the readout weight are adapted in the adjusted in the learning process. 
%

Even though RC methods have been successfully used for solving temporal tasks, the tuning of their parameters can be difficult.
The initialization of the reservoir parameters often requires the human expertise and several empirical trials.
%
%
Over the last years, several approaches have been studied for the reservoir design.
An analysis of the intrinsic plasticity for the ESN model was presented in~\cite{Schrauwen07}.
A specific kind of RC methods uses topographic maps for initializing its weights~\cite{Luko10,BasterCord11, BasterACM13}.
Besides, an evolutionary algorithm was used for designing the reservoir~\cite{Ishu04}. 
%
%
Additionally, other metaheuristic techniques were applied for optimizing the reservoir global parameters, topology and reservoir weights was studied in~\cite{Ferreira10,Ferreira11,Ferreira13}.
%

%
The~\textit{Particle Swarm Optimization (PSO)} is an efficient and widely used metaheuristic for finding optimal regions on complex spaces.
The PSO was applied for defining the spectral radius, the kind of transfer function, the reservoir size and the presence of feedback connections~\cite{Anderson12}.
In this paper, we modify the way of using PSO to construct the reservoir with respect to the approach presented in~\cite{Anderson12}.
We adjust a subset of the reservoir weights, the rest of weights are kept fixed during the training as usual in RC models.
Our hypothesis is that it is enough to tune few weights of the reservoir using the PSO algorithm, in order to improve the ESN performance in terms of computational time and accuracy rate.
This strategy obtains good experimental results, without requiring operations with high computational cost (for instance: it avoids to compute the spectral radius of the reservoir matrix).
This article is structured as follows. 
Section~\ref{Background} presents a background of the two main models used in this work: ESN and PSO.
Section~\ref{Metho} contains the contribution of this work.
Next, we present our experimental results, and then we go for final conclusions and future work.

\section{Background}
\label{Background}
In this Section, we specify the context where the ESN models are applied.
An ESN model is mainly used for solving supervised learning tasks, wherein the data set presents temporal dependencies, although it can be also used for non-temporal supervised learning problems~\cite{Jaeger09}.
Besides, we present a description of both the ESN tool and the PSO technique.

\subsection{Problem Specification}
%
%
Given a training set composed by pairs of discrete-time vectors $(\va(t),\vb(t))$, $\va(t)\in\R^{\Na}$ and $\vb(t)\in\R^{\Nb}$ for all $t$ in an arbitrary interval of time; 
the goal in a supervised learning task is finding a parametric mapping $\phi(\cdot)$ such that a distance function is minimized.
This distance function measures the deviation of the $\phi(\cdot)$ predictions from the target values $\vb$.
Examples of distance functions are the \textit{square error} and the \textit{Kullback-Leibler} distance~\cite{Jaeger09}.
In this article the mapping is given by the ESN model and we evaluate it using the square error distance.

\subsection{Basic Description of the Echo State Network Model}
\label{ESNsection}
The ESN model is a Neural Network composed by a hidden recurrent structure (called \textit{reservoir}) and a readout structure that is a linear regression.
The reservoir role's consists of encoding the temporal information of the input data.
%
%
Besides, the reservoir provides a complex nonlinear transformation of the input patterns, which enhances the linear separability of the input data.
The readout structure is used for supervised training adaptation.
In the canonical ESN tool the readout structure is a linear regression model~\cite{Jaeger09}.
We follow the previous notation concerning the training set.
We use the notation for the components of the ESN model presented in~\cite{Jaeger09}.
The training set is collected in the pairs $(\va(t),\vb(t))$, $t=1,\ldots, T$.
A vector $\x(t)$ represents the reservoir state at each time $t$.
We denote by $\Na,\Nx$ and $\Nb$ the dimensions of the vectors $\va,\x$ and $\vb$, respectively.
%
%
In the canonical ESN, the transfer function of the reservoir neurons is the $\tanh(\cdot)$ function.
The reservoir state is computed as follows:
\begin{equation}
\label{reservoirStateCord}
x_m(t) = \tanh\!\bigg(\!\ds{w_{m0}^{\rm{in}} + \sum_{i=1}^{\Na}w^{\rm{in}}_{mi} a_i(t)} + \ds{\sum_{i=1}^{\Nx}w^{\rm{r}}_{mi} x_i(t-1)}\!\bigg), 
\end{equation}
$\forall m\in[1,\Nx]$, where the weight connections between input and reservoir nodes are given by a~$\Nx\times(\Na+1)$ weight matrix~$\wi$, the connections among the reservoir neurons are represented by a~$\Nx\times\Nx$ weight matrix~$\wre$ and a~$\Nb\times(\Nx+\Na+1)$ weight matrix~$\wo$ represents the connections between reservoir and output units.

The amount of reservoir units is much larger than the dimensionality of the input space ($\Na \ll \Nx$)~\cite{Jaeger09}.
We denote by a vector $\y(t)$ the model output at time $t$, which is generated by a linear regression as follows:
\begin{equation}
\label{regressionOutput}
{y}_m(t) = \ds{w_{m0}^{\rm{out}} + \sum_{i=1}^{\Na} \wmi{out} a_i(t)}  + \ds{\sum_{i=1}^{\Nx}\wmi{out} x_i(t)},
\end{equation}
$\forall m\in[1,\Nb]$. 
%

\subsection{The Particle Swarm Optimization Technique}
The \textit{Particle Swarm Optimization (PSO)} method is an algorithm for finding optimal points on complex search spaces~\cite{Kennedy95}.
The technique is based on social behaviors of a set of particles (\textit{swarm}) in a simplified environment. 
%
The procedure searches for optimal points on a multidimensional space by adjusting vectors that represent particle positions.
The update rule of trajectories is inspired on social interactions.

More formally, let $N$ be the number of particles in the system and $M$ the dimension of the search space.
Each particle~$i$ is characterized by a pair~$(\x_i,\ve_i)$, $\x_i,\ve_i\in\R^M$.
Metaphorically speaking, the vector~$\x_{i}$ represents the position of~$i$ and~$\ve_{i}$ represents the velocity of $i$.
We denote by~$\p_{i}(t)$ the best position of~$i$ ever found at time~$t$.
Let~$\p^*(t)$ be a vector with the information of the best swarm position that has ever found until time~$t$.
The algorithm is iterative, at each epoch the objective function (function to be optimized) is evaluated, next the vectors~$\x_i$ and~$\ve_i$ are updated for all~$i$.
At any time~$t$, the system dynamics are given by the expressions~\cite{Clerc02}:
\begin{equation}
\label{velocity}
\ve_{i}(t+1) = \iota\ve_{i}(t) + \delta^{1}_{i}(t) \big(\p_{i}(t) - \x_{i}(t)\big) + \delta^{2}_{i}(t) \big(\p^*(t) - \x_{i}(t)\big),
\end{equation}
and
\begin{equation}
\label{position}
\x_{i}(t+1)=\x_{i}(t)+\ve_{i}(t+1),
\end{equation}
where the parameter~$\iota\in(0,1)$ is called the \textit{inertia},~$\delta^{1}$ and~$\delta^{2}$ are two diagonal matrices. 
The inertia controls the tradeoff between exploitation and exploration on the search space.
The diagonal elements~$\delta^1_{i}$ and~$\delta^2_{i}$ are uniformly distributed in~$[0,\varphi_1]$ and~$[0,\varphi_2]$, respectively.
These matrices weight the relationship between individual positions and the ``good'' local and global position. 
For this reason, the parameters~$\varphi_1$ and~$\varphi_2$ are called the \textit{acceleration coefficients}.
A pseudo-code of the PSO technique is presented in Algorithm~\ref{AlgoPSO}.
\begin{algorithm}
\caption{Specification of the Particle Swarm Optimization used for finding the weight matrix of the reservoir.}
\label{AlgoPSO}
\begin{algorithmic}
	%
	\STATE $t=t_0$;
	\STATE Initialize population~$(\x_{i},\ve_i)(t), \forall i$;
	\STATE Evaluate~$F(\x_i), \forall i$;
	\STATE Set~$\p^*(t)$ and~$\p_{i}(t)$ for all~$i$;
	\WHILE {(termination criterion is not satisfied)}
		\FOR{(each particle~$i$)}
			\STATE Compute~$\ve_{i}(t+1)$ using~(\ref{velocity});
			\STATE Compute~$\x_{i}(t+1)$ using~(\ref{position});
			\STATE Evaluate~$F(\x_i)$;
			\STATE Update local best~$\p_{i}(t+1)$;
		\ENDFOR
		\STATE Update global best~$\p^*(t+1)$;
		\STATE $t=t+1$;
	\ENDWHILE
	\STATE Return $\p^*(t)$;
	\end{algorithmic}
\end{algorithm}

\section{The PSO for Setting the ESN Model}
\label{Metho}
The performance of the ESN model basically depends of the following global parameters: the input scaling factor, the reservoir size, the spectral radius of the reservoir matrix and the topology of the reservoir network.
The input scaling factor controls the impact of the inputs over the reservoir state~\cite{Butcher2013}. 
In the RC literature has been used a large sparse pool of interconnected neurons in the reservoir.
A reservoir projection in a larger space improves the model accuracy, although there is a tradeoff to reach in the reservoir size. 
A too large reservoir can provoke the over-fitting phenomenon.
The spectral radius impacts on the stability and chaoticity of the reservoir dynamics, as a consequence it influences on the memory capability of the model. 
The stability of the ESN reservoir is guaranteed when the spectral radius is less than~$1$, this stability condition was established in the~\textit{Echo State Property (ESP)}~\cite{Jaeger01}.
According to previous experiences, it has not been clear what the impact of the reservoir density would be on the model accuracy.
Although, sparse matrices process the information faster than dense matrices, as a consequence a sparse reservoir can improve performance in time~\cite{Jaeger09,MantasPracticalGuide12}. 
%
%
Recently, an evolutionary algorithm was used to find the reservoir size, the spectral radius and the density of the reservoir matrix~\cite{Ishu04}. 
In addition, evolutionary and genetic algorithms were applied for optimizing the reservoir global parameters and for designing the connectivity of the reservoir~\cite{Ferreira10,Ferreira11,Ferreira13}.
%
%

The PSO technique was already used for defining the spectral radius and other main parameters of the reservoir in~\cite{Anderson12}.
Nevertheless, it is known that different reservoirs with the same spectral radius can have a substantial variance in the model accuracy~\cite{Schrauwen07}.
In recurrent topologies, to compute the eigenvalues modulus can be not-robust and computational expensive.
The converge rate of the spectrum computation is determined by how close certain eigenvalues are to zero.
Besides, the operation of rescaling the reservoir matrix by the spectral radius has a high computational cost~\cite{Ferreira13}. 
%
%

%
In this article, we propose a hybrid method which uses the PSO for adjusting a subset of the reservoir weights without requiring to compute the spectrum of the reservoir matrix.
We do not use the PSO for finding the spectral radius, and the other global parameters.
%

\begin{figure}[h]
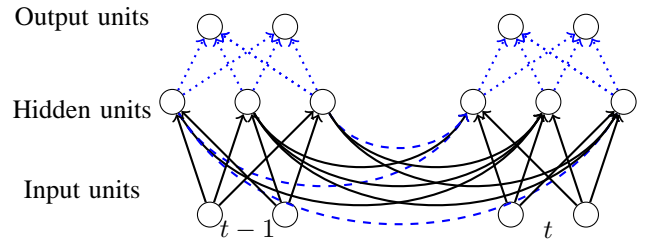

\vspace{-0.2in}
\begin{center}
{\tikz[every node/.style={draw,circle}]{
{\node[fill=white]  (a) at (-6.2,1.1)[fill=white,draw=white,text=black] {Output units};}
{\node[fill=white]  (a) at (-6.2,-0.1)[fill=white,draw=white,text=black] {Hidden units};} 
{\node[fill=white]  (a) at (-6.2,-1.2)[fill=white,draw=white,text=black] {Input units};}
{\node[fill=white] (t1) at (-4,-1.7)[fill=white,draw=white,text=black] {$t-1$};}
{\node[fill=white] (t2) at (0,-1.7)[fill=white,draw=white,text=black] {$t$};}
{\node[fill=white] (ha1) at (-5,0) {};}
{\node[fill=white] (ha2) at (-4,0) {};}
{\node[fill=white] (ha3) at (-3,0) {};}
{\node[fill=white] (hb1) at (-1,0) {};}
{\node[fill=white] (hb2) at (0,0) {};}
{\node[fill=white] (hb3) at (1,0) {};}
{\node[fill=white] (ia1) at (-4.5,-1.5) {};}
{\node[fill=white] (ia2) at (-3.5,-1.5) {};}
{\node[fill=white] (ib1) at (-0.5,-1.5) {};}
{\node[fill=white] (ib2) at (0.5,-1.5) {};}
{\node[fill=white] (oa1) at (-4.5,1) {};}
{\node[fill=white] (oa2) at (-3.5,1) {};}
{\node[fill=white] (ob1) at (-0.5,1) {};}
{\node[fill=white] (ob2) at (0.5,1) {};}
\path[->,dashed,blue,every loop/.style={looseness=5}] (ha1)  edge[dashed,->,thick,in=240,out=-60] (hb1);
\path[->,every loop/.style={looseness=5}] (ha1)  edge[->,thick,in=240,out=-60] (hb2);
\path[->,dashed,blue,every loop/.style={looseness=5}] (ha1)  edge[dashed,->,thick,in=240,out=-60] (hb3);
\path[->,every loop/.style={looseness=5}] (ha2)  edge[->,thick,in=240,out=-60] (hb1);
\path[->,every loop/.style={looseness=5}] (ha2)  edge[->,thick,in=240,out=-60] (hb2);
\path[->,every loop/.style={looseness=5}] (ha2)  edge[->,thick,in=240,out=-60] (hb3);
\path[->,dashed,blue,every loop/.style={looseness=5}] (ha3)  edge[dashed,->,thick,in=240,out=-60] (hb1);
\path[->,every loop/.style={looseness=5}] (ha3)  edge[->,thick,in=240,out=-60] (hb2);
\path[->,every loop/.style={looseness=5}] (ha3)  edge[->,thick,in=240,out=-60] (hb3);
{\draw (ha1)  edge[blue,dotted,->,thick] (oa1);}
{\draw (ha1)  edge[blue,dotted,->,thick] (oa2);}
{\draw (ha2)  edge[blue,dotted,->,thick] (oa1);}
{\draw (ha2)  edge[blue,dotted,->,thick] (oa2);}
{\draw (ha3)  edge[blue,dotted,->,thick] (oa1);}
{\draw (ha3)  edge[blue,dotted,->,thick] (oa2);}
{\draw (ha3)  edge[blue,dotted,->,thick] (oa1);}
{\draw (hb1)  edge[blue,dotted,->,thick] (ob1);}
{\draw (hb1)  edge[blue,dotted,->,thick] (ob2);}
{\draw (hb2)  edge[blue,dotted,->,thick] (ob1);}
{\draw (hb2)  edge[blue,dotted,->,thick] (ob2);}
{\draw (hb3)  edge[blue,dotted,->,thick] (ob1);}
{\draw (hb3)  edge[blue,dotted,->,thick] (ob2);}
{\draw (hb3)  edge[blue,dotted,->,thick] (ob1);}
{\draw (ia1)  edge[->,thick] (ha1);}
{\draw (ia1)  edge[->,thick] (ha2);}
{\draw (ia1)  edge[->,thick] (ha3);}
{\draw (ia2)  edge[->,thick] (ha1);}
{\draw (ia2)  edge[->,thick] (ha2);}
{\draw (ia2)  edge[->,thick] (ha3);}
{\draw (ib1)  edge[->,thick] (hb1);}
{\draw (ib1)  edge[->,thick] (hb2);}
{\draw (ib1)  edge[->,thick] (hb3);}
{\draw (ib2)  edge[->,thick] (hb1);}
{\draw (ib2)  edge[->,thick] (hb2);}
{\draw (ib2)  edge[->,thick] (hb3);}
}
}
\vspace{-0.2in}
\caption{\label{RecNetwork}An example of the topology of the PSO-ESN model. A solid black line represents fixed weight during the learning process, blue dashed lines represent the weights adjusted by the PSO, and blue dotted lines represent the readout weights adjusted using a memoryless supervised learning method (for instance: linear regression model).}
\end{center}
\vspace{-0.1in}
\end{figure}

The weights can be classified into the following categories: input weights, random reservoir weights, reservoir weights adjusted by PSO and the readout weights.
We denote by~$\Omega^{\rm{in}}$ the set of input weights that are collected in the matrix~$\wi$, we denote by~$\Omega^{\rm{r}}$ the reservoir weights that are collected in the matrix~$\wre$, and we denote by~$\Omega^{\rm{out}}$ the readout weights collected in the matrix~$\wo$.
Let~$\Omega^{\rm{h}}$ be the subset of the reservoir weights ($\Omega^{\rm{h}}\subset\Omega^{\rm{r}}$) that are adjusted using the PSO method.
The weights in~$\Omega^{\rm{h}}$ are hidden weights randomly selected from~$\Omega^{\rm{r}}$.
The relationship between the cardinality of~$\Omega^{\rm{h}}$ and~$\Omega^{\rm{r}}$ is given by~$\vert\Omega^{\rm{h}}\vert=\alpha\vert\Omega^{\rm{r}}\vert$ where~$\alpha\in (0,1)$ and~$\vert\cdot\vert$ is the cardinality function of a set. The parameter~$\alpha$ is empirically estimated. 
Figure~\ref{RecNetwork} presents an example of the different kind of parameters, wherein $\Omega^{\rm{h}}$ and $\Omega^{\rm{out}}$ are represented by blue dashed and dotted lines, respectively. Other weights are represented by black solid lines. Only the blue weights are adjusted in this approach. 
In summary, the procedure to train this hybrid model is presented in~\ref{AlgoHybrid}.
%

\begin{algorithm}
\caption{Pseudo-algorithm of the PSO-based phase for setting the ESN model.}
\label{AlgoHybrid}
\begin{algorithmic}
	%
	\STATE Initialize the PSO parameters: $M$, $N$, $\iota$, $\delta^1$, $\delta^2$;
	\STATE Initialize the weights~$\Omega^{\rm{in}}$ and~$\Omega^{\rm{r}}$ using a random distribution;
	\STATE Select $\Omega^{\rm{h}}\subset \Omega^{\rm{r}}$ using a random distribution;
	\REPEAT 
	{
	\STATE Apply the PSO for optimizing~$\Omega^{\rm{h}}$ (Algo.~\ref{AlgoPSO});
	\STATE Compute the~$\wo$ using linear ridge regression;
	\STATE Evaluate the accuracy of the model;
	}
	\UNTIL{criterion is satisfied}
	\STATE Return the network weights;
	\end{algorithmic}
\end{algorithm}
\vspace{-0.2in}


%
\section{Empirical Results}
\label{EmpiricalResults}
In this section, we provide the performance of the canonical ESN model and the hybrid method introduced in the precedent section on four benchmark experiments.
We use the acronym \textit{PSO-ESN} for denoting the procedure proposed in this work.
%
We call \textit{epoch} to an iteration of the training algorithm through all the examples in the training set.
In order to have statistically significant results, we run each model on each benchmark using~$30$ different random initializations.
In the case of the PSO algorithm, for each benchmark test we use a grid points of values~$M$ and~$N$.

%

%
We compare the following procedures:
\begin{itemize}
\item ESN: we initialize the network weights using an Uniform random distribution~$U[w_{min},w_{max}]$. 
The topology consists of a network with three fully connected layers (input, reservoir and output layer).
We control the density of the reservoir and the spectral radius module.
We rescale the weights of~$\wre$ using the spectral radius in order to ensure the ESP. 
We project the input space using the reservoir. 
Next, we compute the readout weights using the training set and standard ridge regression.
We repeat the experiment evaluating the performance for several spectral radius of~$\wre$ values. In our experiments, the reservoir size and density are fixed.
\item PSO-ESN: we initialize the network weights~$\Omega^{\rm{in}}$ and~$\Omega^{\rm{r}}$ using a uniform random distribution~$U[w_{min},w_{max}]$. 
Next, we random select a subset~$\Omega^{\rm{h}}$ such that~$\Omega^{\rm{h}}\subset\Omega^{\rm{r}}$.
Then, we apply the PSO for setting the weights in~$\Omega^{\rm{h}}$.
In this step we consider the \textit{Mean Square Error (MSE)} as fitness function in the PSO algorithm.  
Finally, we use the training set for computing the readout weights.
\end{itemize}
The statistical comparison between the accuracy reached by the two methods was realized using confidence intervals.
We use asymptotic confidence intervals of the mean of the accuracy reached on the different experiments.

The remains of this section includes a description of the data set, the experimental setting and the reached results.
\subsection{Description of the Benchmarks}
We use the following range of benchmark problems.
The first data set is an experimental data measured with a \textit{LeCroy} oscilloscope, the patterns corresponds to the intensity pulsations of a laser.
This benchmark is often called as the \textit{Santa Fe Laser} data.
The data is a cross-cut through periodic to chaotic intensity laser pulsations, which more or less follow the theoretical Lorenz model of a two level system~\cite{SantaFeLaser}. 
The task consists to predict the next laser pulsation~$b(t+1)$, given the precedent values up to~$t$.
The original data only consists of~$1000$ measurements, we use~$499$ for training and~$500$ for test. We use a washout of~$30$ samples.
The initial input weights are in~$[-0.8,0.8]$ and the initial reservoir weights are in~$[-0.2,0.2]$.
The regularization parameter ($\gamma$) used for computing the readouts was set with~$0.001$.
The reservoir size has~$50$ units, the spectral radius and the sparsity of the reservoir matrix were~$0.9$ and~$0.3$, respectively. 

The Nonlinear Autoregressive Moving Average (NARMA) is a widely studied benchmark problem~\cite{Rodan11,Jaeger01,Verstraeten07,Ferreira13}.
The interests of this data is based on the high degree of chaos in its dynamics. Additionally, the data can present long-range dependency, as a consequence to learn patterns on the training set is a difficult task~\cite{Bengio94}.  
%
The sequence of patterns is generated by the expression:
\beq 
b(t+1)=c_1 b(t)+c_2 b(t)\displaystyle{\sum_{i=0}^{k-1}b(t-i)}+c_3 s(t-(k-1))s(t)+c_4,
\eeq
where~$s(t)\sim U[0,0.5]$ and the constants values are~$c_1=0.3$,~$c_2=0.05$,~$c_3=1.5$ and~$c_4=0.1$.
The data set was rescaled in~$[0,1]$.
In order to evaluate the memory capability of the model, we consider two simulated NARMA series with~$k=10$ and~$k=30$.
For the case of~$10$th order NARMA, we generate a training data with~$1990$ samples and a test set with~$390$ samples.
The~$30$th order NARMA training set has~$2772$ samples and the test set has~$1428$ patterns.
The~$70\%$ of the weight connections among reservoir units are zeros.
The reservoir size is~$150$ units for the~$10$th order NARMA and~$200$ units for the another NARMA benchmark.

The last benchmark problem refers to the traffic prediction on the Internet.
The data is from an Internet Service Provider (ISP) working in~$11$ European cities. 
The original data was collected in bits using a time interval of five minutes.  
The size of the training and test data set are~$9848$ and~$4924$.
%
%
The goal is to predict the Internet traffic at time~$t+1$ using the information from~$t-6$ up to time~$t$. 
More details about this data set and a forecasting analysis can be seen in~\cite{Baster12ESQN,Cortez12,BasterIBICA14}.

\subsection{First Results}
Table~\ref{Res} summarizes the accuracy reached by the PSO-ESN on the experiments.
First column identifies the benchmark task and second column refers to the dimension of each particle in the PSO technique.
Last two columns indicate the performance of the PSO-ESN.
Third column is the MSE average performed on~$30$ different initializations and fourth column is the standard deviation of the MSE computed on the different initializations.

Table~\ref{Res_Classic} presents the performance of the ESN model. Second column shows the mean of accuracy reached on the $30$ trials and the third columns refers to the standard deviation of this error measures.
Table~\ref{Res_Traffic} shows the accuracy reached for both models on the training and test Internet traffic data set.
The real values in the tables are written using the scientific notation form.

%
We can generate a confidence interval (CI) of the MSE~$[e_{min},e_{max}]$ using the standard deviation of the tables~\ref{Res} and~\ref{Res_Classic}.
Let~$[e^{1}_{min},e^{1}_{max}]$ be the CI for the MSE obtained with the PSO-ESN method, and let~$[e^{2}_{min},e^{2}_{max}]$ be the CI computed for the MSE reached for the ESN model.
Note that, if we generate~$95\%$ CI considering an approximation normal distribution, then~$[e^{1}_{min},e^{1}_{max}]$ and~$[e^{2}_{min},e^{2}_{max}]$ are distinct intervals.
Specifically, we have~$e^{1}_{max}<e^{2}_{min}$ for the four benchmarks studied in this work.

\begin{figure}[h!t]
\begin{center}
\includegraphics[angle=0,width=0.5\textwidth]{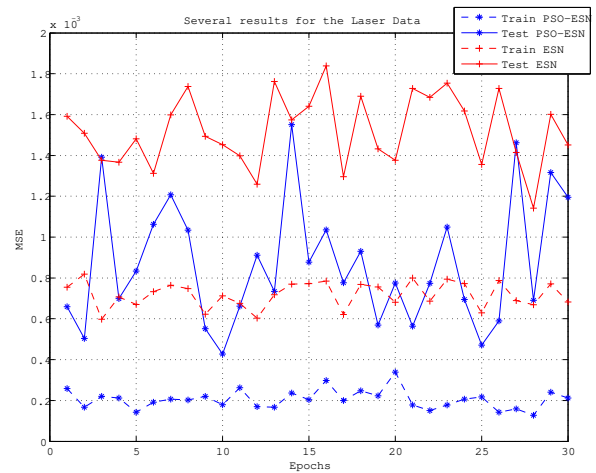}
\caption{Accuracy of several initializations of the PSO-ESN and ESN for the train and test set of Laser data set. 
The reservoir has $50$ units, $M=5$ and the PSO has $10$ particles.
In spite that the epochs are independent of each other, for a better visualisation we draw a continuos curve for the testing experiment and dashed curves for the training experiments.
}
\label{Laser}
\end{center}
\vspace{-0.25in}
\end{figure}

Figure~\ref{Laser} shows the different accuracy reached for both models with the Laser data set.
Red lines corresponds to the ESN model and blue lines refers to the PSO-ESN.
The figure shows the error obtained with the training and test data set versus different initializations. 

\begin{figure}[h!t]
\begin{center}
\includegraphics[angle=0,width=0.5\textwidth]{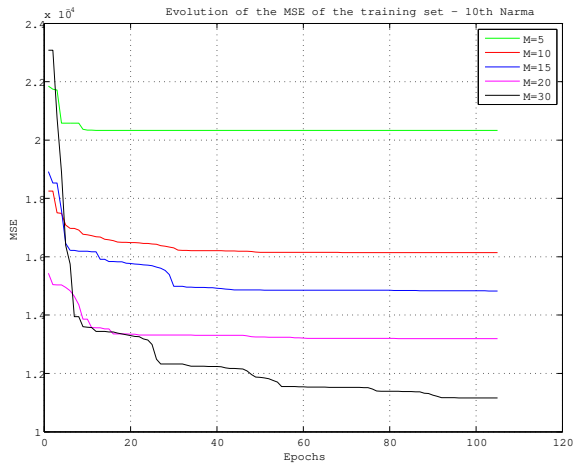}
\caption{Example of the evolution of the MSE of the training set according to $M$ for the $10$th order NARMA data set. 
The MSE is for one specific random initialization of the weights. The reservoir has $150$ units and the PSO has $35$ particles.}
\label{EvolMTrain}
\vspace{-0in}
\end{center}
\end{figure}

Figure~\ref{EvolMTrain} illustrates the influence of the parameter $M$ on the accuracy of the PSO-ESN model.
This parameter represents the dimension of each particle of the swarm, this means the reservoir weights that are adjusted using the PSO.
According to the figure, we can see that larger $M$ values reached better accuracy.
For instance, in the Figure~\ref{EvolMTrain} for a number of epochs equal to $60$ the line at the top corresponds to $M=5$ and the line at the bottom corresponds to $M=30$ (the order of lines from top to bottom is $M=5,10,15,20$ and $30$).
On the other hand, a larger search space (larger value of $M$) can increase the running time and can cause the over-fitting phenomenon.
According our empirical results, it is enough to have $M=\Nx/5$ to have better accuracy than the ESN model rescaling the reservoir weights with the spectral radius.

\begin{table}[h!t]
\begin{center}
\caption{\label{Res} Performance of the  PSO-ESN hybrid method. 
Performance of the test data set for the Laser and NARMA benchmark problems. 
The second column corresponds to the dimension of each particle in the PSO algorithm. The columns~$3$ and~$4$ are the average and the standard deviation of the accuracy obtained for the $30$ experiments. 
}
\begin{tabular}{*{4}{c}}\hline\hline
Data set & M & Mean & Stdv  \\
\hline
Laser data & $5$ & $8.6657\times 10^{-4}$ & $9.3541\times 10^{-9}$\\
$10$th NARMA & $15$ & $2.0462\times 10^{-4}$ & $1.9416\times 10^{-9}$\\
$10$th NARMA & $30$ & $1.9623\times 10^{-4}$ & $1.4595\times 10^{-9}$\\
$30$th NARMA & $40$ & $1.3247\times 10^{-2}$ & $1.5476\times 10^{-7}$\\\hline\hline
\end{tabular}
\end{center}
\vspace{-0.2in}
\end{table}
\begin{table}[h!t]
\begin{center}
\caption{
\label{Res_Classic} 
Performance of the ESN model.
Performance of the test data set for the Laser and NARMA benchmark problems. The columns~$3$ and~$4$ are the average and the standard deviation of the accuracy obtained for the $30$ experiments.
}
\begin{tabular}{*{3}{c}}\hline\hline
Data set & Mean & Stdv  \\
\hline
Laser data & $1.5220\times 10^{-3}$& $3.0711\times 10^{-8}$\\
$10$th NARMA  & $2.0538\times 10^{-4}$ & $1.6871\times 10^{-9}$\\
$30$th NARMA  & $1.4025\times 10^{-2}$ & $1.2221\times 10^{-7}$\\\hline\hline
\end{tabular}
\end{center}
\vspace{-0.05in}
\end{table}
\begin{table}[h!t]
\begin{center}
\caption{
\label{Res_Traffic}
Performance of both PSO-ESN and ESN for the train and test data set for the Internet traffic prediction.
The columns~$3$ and~$4$ are the average and the standard deviation of the accuracy obtained for the $30$ experiments.
}
\begin{tabular}{*{4}{c}}\hline\hline
Method & Data set & Mean & Stdv  \\
\hline
PSO-ESN & Train & $7.1613 \times 10^{-8}$ & $ 1.6245\times 10^{-15}$\\
ESN & Train & $1.2802\times 10^{-7}$ & $1.6833\times 10^{-14}$\\
PSO-ESN &Test & $1.5293\times 10^{-6}$ & $ 1.3340\times 10^{-12}$\\
ESN &Test & $2.5661\times 10^{-6}$ & $3.6516\times 10^{-12}$\\\hline\hline
\end{tabular}
\end{center}
\vspace{-0.2in}
\end{table}

%
%

\section{Conclusions and Future Work}
%
%
%

%
%
In this article we present a method that uses the \textit{Particle Swarm Optimization (PSO)} for initialization of the \textit{Echo State Networks (ESN)} is proposed for solving temporal supervised learning tasks. 
The ESN model is an efficient technique to train and design a Recurrent Neural Network.
On the other hand, the PSO algorithm has been successfully used for optimizing continuous functions.  
%

Over the last years, several approaches have been presented for designing the reservoir.
In this contribution, we use the PSO for adjusting a subset of the reservoir weights.
To tune all the reservoir weights using meta-heuristics can be a very expensive task.
As a consequence, a subset of the reservoir weights is randomly selected and adjusted using the PSO.
The setting of the reservoir weights is realized in an automatic way using the PSO.
%
Besides, the procedure does not require to compute the spectrum of the reservoir matrix, which is a computational expensive operation.

As a for future work, we can extend the same procedure to other Reservoir Computing methods.
As well as, we are interesting in comparing the performance reached by the PSO algorithm with other bio-inspired techniques.

\bibliographystyle{IEEEtran}
\bibliography{IEEEabrv,refRnn}
%

%

\end{document}